\documentclass[runningheads]{llncs}
\usepackage{graphicx}
\usepackage{cite}
\usepackage{graphicx}
\usepackage{amsmath}
\usepackage{amssymb}
\usepackage{dsfont}
\usepackage{graphicx,subcaption}
\usepackage{booktabs}
\usepackage{algorithm}
\usepackage{algpseudocode}
\usepackage{algcompatible}
\usepackage{multirow}
\usepackage{booktabs}
\usepackage{mathtools}
\usepackage{makecell}
\usepackage{xcolor}
\usepackage[normalem]{ulem}
\usepackage{wrapfig}

\usepackage[colorlinks=true,allcolors=blue]{hyperref}
\begin{document}
%
\title{Contrastive Transformer-based Multiple Instance Learning for Weakly Supervised Polyp Frame Detection}
%
%
\author{
Yu Tian \inst{1,2},
Guansong Pang \inst{3},
Fengbei Liu \inst{1}, 
Yuyuan Liu \inst{1}, 
Chong Wang \inst{1},  
Yuanhong Chen \inst{1},   
Johan W Verjans \inst{1,2}, and 
Gustavo Carneiro \inst{1}
}

%

\institute{Australian Institute for Machine Learning, University of Adelaide
\and
South Australian Health and Medical Research Institute
\and
 Singapore Management University
    }
\maketitle              
\begin{abstract}
Current polyp detection methods from colonoscopy videos use exclusively normal (i.e., healthy) training images, which
i) ignore the importance of temporal information in consecutive video frames, and ii) lack knowledge about the polyps. 
Consequently, they often have high detection errors, especially on challenging polyp cases (e.g., small, flat, or partially visible polyps).
In this work, 
we formulate polyp detection as a weakly-supervised anomaly detection task that uses  video-level labelled training data to detect frame-level polyps. 
In particular, we propose a novel convolutional transformer-based multiple instance learning method designed to identify abnormal frames (i.e., frames with polyps) 
from anomalous videos (i.e., videos containing at least one frame with polyp).
In our method, local and global temporal dependencies are seamlessly captured while we simultaneously optimise video and snippet-level anomaly scores. 
A contrastive snippet mining method is also proposed to enable an effective modelling of the challenging polyp cases.
The resulting method achieves a detection accuracy that is substantially better than current state-of-the-art approaches on a new large-scale colonoscopy video dataset introduced in this work. 
Our code and dataset will be publicly available upon acceptance. 

\end{abstract}

\section{Introduction and Background}
\label{sec:introduction}

\indent Colonoscopy has become a vital exam for colorectal cancer (CRC) early diagnosis. 
This exam targets the early detection of  polyps (a precursor of colon cancer), which can improve survival rate by up to 95\%~\cite{ji2021progressively,tian2020few,pu2020computer,tian2021detecting,tian2019one}.
During the procedure, doctors inspect
the lower bowel with a scope to find polyps, but the quality of the exam depends on the ability of doctors to avoid mis-detections~\cite{pu2020computer}. 
This can be alleviated by systems that automatically assist doctors detect frames containing polyps from colonoscopy videos. 
Nevertheless, accurate polyp detection is challenging due to the variable appearance, size and shape of colon polyps and their rare occurrence in an colonoscopy video.

One way to mitigate polyp detection challenges is with fully supervised training approaches, but given the expensive acquisition of fully labelled training sets, recent approaches have formulated the problem as an unsupervised anomaly detection (UAD) task~\cite{liu2019photoshopping,tian2021constrained,chen2021deep,tian2021self}.
These UAD methods~\cite{liu2019photoshopping,tian2021constrained,chen2021deep,tian2021self} are trained with only normal training images and videos, and abnormal testing images and videos that contain polyps are detected as anomalous events.
However, UAD approaches do not use training images or \textit{snippets} (i.e., a set of consecutive video frames) containing polyps, so they are ineffective in recognising polyps of diverse characteristics, especially those that are small, partially visible, or irregularly shaped.
As shown in a number of recent studies~\cite{tian2020few,tian2021weakly,Wu2020not,sultani2018real,pang2019deep,pang2021toward}, incorporating some knowledge about anomalies into the training of anomaly detectors has improved the detection accuracy of hard anomalies.
For example, weakly-supervised video anomaly detection (WVAD)~\cite{tian2021weakly,Wu2020not,sultani2018real} relies on video-level labelled data to train detection models.
The video-level labels only indicate whether the whole video contains anomalies or not, which is easier to acquire than fully-labelled datasets with frame-level annotations. The WVAD formulation is yet to be explored in the detection of polyps from colonoscopy, but it is of utmost importance because colonoscopy videos are often annotated with video-level labels in real-world datasets.

Most existing WVAD methods~\cite{tian2021weakly,Wu2020not,sultani2018real,zhong2019graph,feng2021mist} rely on multiple instance learning (MIL), in which all snippets in a normal video are treated as normal snippets, while each abnormal video is assumed to have at least one abnormal snippet.
This approach can utilise video-level labels to train an anomaly-informed detector to find anomalous frames, but MIL methods often fail to select rare abnormal snippets in  anomalous videos, especially the challenging abnormal snippets that have subtle visual appearance differences from the normal ones (e.g., small and flat colon polyps or frames with partially visible polyps--see Fig.~\ref{fig:hard_abn_mining}). Consequently, they perform poorly in detecting these subtle anomalous snippets.
Moreover, the WVAD methods above are trained on 
individual images, ignoring the important temporal dependencies in colonoscopy videos that 
can be explored for a more stable polyp detection performance.

In this paper, we 
introduce the first WVAD method specifically designed for detecting polyp frames from colonoscopy videos.
Our method introduces a new contrastive snippet mining (CSM) algorithm to identify hard and easy normal and abnormal snippets. These snippets are further used to simultaneously optimise video and snippet-level anomaly scores, which effectively reduces detection errors, such as mis-classifying snippets with subtle polyps as normal ones, or normal snippets containing feces and water as abnormal ones.
The exploration of global temporal dependency 
is also incorporated into our model with a transformer module, enabling a more stable anomaly classifier for colonoscopy videos. 
To resolve the poor modelling of local temporal dependency suffered by the transformer module~\cite{wu2021cvt}, we also propose a convolutional transformer block to capture local correlations between neighbouring snippets.   
Our contributions are summarised as follows:
\begin{itemize}
\item To the best of our knowledge, this is the first work to tackle polyp detection from colonoscopy in a weakly supervised video anomaly detection manner. 
\item We propose a new transformer-based MIL framework that optimises anomaly scores in both snippet and video levels, resulting in more accurate anomaly scoring of polyp snippets.
\item We introduce a new contrastive snippet mining (CSM) approach to identify hard and easy normal and abnormal snippets, 
where we pull the hard and easy snippets of the same class (i.e., normal or abnormal) together using a contrastive loss. This helps improve the robustness in detecting subtle polyp tissues and challenging normal snippets containing feces and water. 
\item We propose a new WVAD benchmark containing a large-scale diverse colonoscopy video dataset that combines several public colonoscopy datasets. 
\end{itemize}
Our extensive empirical results show that our method achieves substantially better results than six state-of-the-art (SOTA) competing approaches on our newly proposed benchmark. 

\section{Method}

Our method is trained with a set of weakly-labelled videos $\mathcal{D} = \{ (\mathbf{F}_i,y_i) \}_{i=1}^{|\mathcal{D}|}$, where $\mathbf{F} \in \mathcal{F} \subset \mathbb{R}^{T \times D}$ 
represents pre-computed features (e.g., I3D~\cite{carreira2017quo}) of dimension $D$ from $T$ video snippets, and $y \in \mathcal{Y} = \{0,1\}$ denotes the video-level annotation ($y_i=0$ if $\mathbf{F}_i$ is a normal video and $y_i=1$ otherwise), with each video being equally divided into a fixed number of snippets.
Our method aims to learn a convolutional transformer MIL anomaly classifier for the $T$ snippets, as in
$r_{\theta,\phi}:\mathcal{F} \to [0,1]^{T}$, where this function is decomposed as
$r_{\theta,\phi}(\mathbf{F})=s_{\phi}(f_{\theta}(\mathbf{F}))$, with
$f_{\theta}:\mathcal{F} \to \mathcal{X}$ being the transformer-based temporal feature encoder parameterised by $\theta$ (with $\mathcal{X} \subset \mathbb{R}^{T \times D}$) and
$s_{\phi}:\mathcal{X} \to [0,1]^T$
denoting the MIL anomaly classifier, parameterised by $\phi$, to optimise snippet-level anomaly scores.

\subsection{Convolutional Transformer MIL Network}

\begin{figure}
    \centering
    \includegraphics[width=0.95\textwidth]{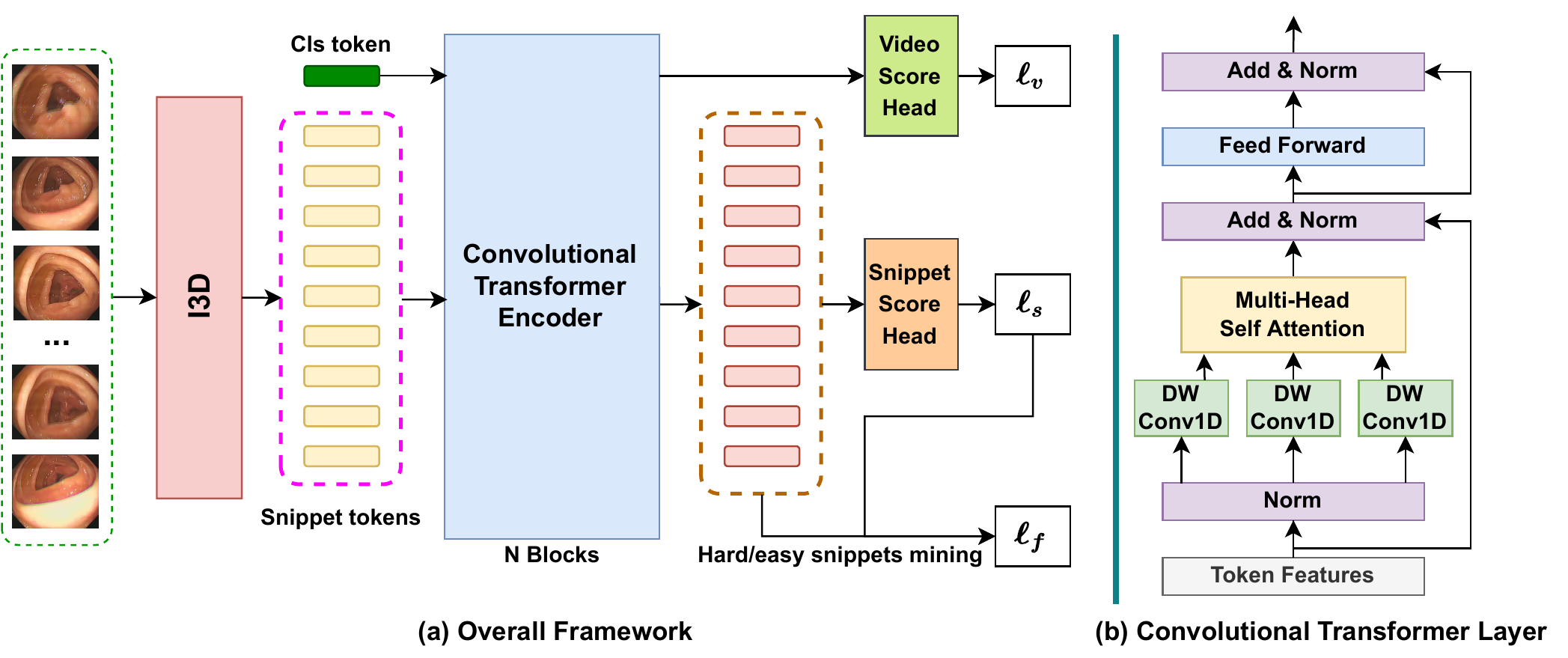}
    \caption{(a) The architecture of our method consists an I3D~\cite{carreira2017quo} snippet feature extractor and a Convolutional Transformer MIL Network. The I3D features are considered as snippet feature tokens to the transformer to predict snippet-wise anomaly scores using a snippet classifier. 
    The Cls token is applied for a video classifier to predict if a video contains anomalies. 
    The output features from the transformer are utilised to mine hard and easy snippets from normal and abnormal videos. 
    The anomaly scores and hard/easy snippet representations are optimised by three proposed losses in~\eqref{eq:main_loss}.  (b) The proposed Temporal Convolutional Transformer Layer replaces the linear projection with depthwise separable convolution (DW Conv1D)~\cite{chollet2017xception}.}
    \label{fig:framework}
    \vspace{-10pt}
\end{figure}

Motivated by the recent success of transformer architectures in analysing the global context of images~\cite{dosovitskiy2020image} and videos~\cite{arnab2021vivit}, 
we propose to use a transformer 
to model the temporal information between the snippets of colonoscopy videos. 
Standard transformer without convolution~\cite{dosovitskiy2020image} cannot learn the local structure between adjacent snippets, which is important for modelling local temporal relations because adjacent snippets are often highly correlated~\cite{tian2021constrained,sultani2018real,Wu2020not}. 
Hence, we replace the linear token projection of the transformer by convolution operations. 
More specifically, we follow~\cite{wu2021cvt} and adopt the depth-wise separable 1D  convolution~\cite{chollet2017xception} on the temporal dimension, as shown in Fig.~\ref{fig:framework}(b).
As shown in Fig.~\ref{fig:framework}(a), the encoder comprises $N$ convolutional transformer blocks 
that produce the final temporal feature representation $\mathbf{X} = f_{\theta}(\mathbf{F})$.

\subsection{Transformer-based MIL Training}

The training of our model comprises a joint optimisation of a transformer-based temporal feature learning, a contrastive snippet mining (CSM) that is used to train a CSM-enabled MIL classifier, and a video-level classifier, with 
\begin{equation}
    \theta^*, \phi^*, \gamma^* = \arg\min_{\theta, \phi, \gamma}
    \ell_{cnt}(\mathcal{D};\theta) + 
    \ell_{snp}(\mathcal{D};\theta,\phi) + \ell_{vid}(\mathcal{D};\theta,\gamma) + \ell_{reg}(\mathcal{D};\theta,\phi)
   \label{eq:main_loss} 
\end{equation}
where 
$\ell_{cnt}(.)$ denotes a contrastive loss that uses the mined hard and easy normal and abnormal snippet features, $\ell_{snp}(.)$ is a loss function to train the snippet classifier $s_{\phi}(.)$ using the top $k$ snippet-level anomaly scores from normal and abnormal videos, $\ell_{vid}(.)$ is a loss function to train the video classifier to predict whether the video contains anomalies, $\theta$, $\phi$ and $\gamma$ are respectively parameters of $\ell_{cnt}(.)$, $\ell_{snp}(.)$ and $\ell_{vid}(.)$, and the regularisation loss is defined by
\begin{equation}
\ell_{reg}(\mathcal{D};\theta,\phi)=
\sum_{(\mathbf{F}_i,y_i)\in\mathcal{D}}
\alpha \left ( \dfrac{1}{T} \sum_{t = 2}^T (\tilde{y}_{i}(t) - \tilde{y}_{i}(t-1))^2 \right )  + \beta \left ( \dfrac{1}{T}
 \sum_{t = 1}^T |\tilde{y}_i(t)| \right ),
    \label{eq:loss_regularisation}
\end{equation}
with $\tilde{y}_{i}(t)\in[0,1]$ denoting the anomaly classifier output for the $t^{th}$ snippet from
$\tilde{y}_i=s_{\phi}(f_{\theta}(\mathbf{F}_i))$. Note that in~\eqref{eq:loss_regularisation}, the first term is a temporal smoothness regularisation, given that anomalous and normal events tend to be temporally consistent~\cite{sultani2018real}, the second term is the sparsity regularisation formulated based on the assumption that anomalous snippets are rare events in abnormal videos, and $\alpha$ and $\beta$ are the hyper-parameters that weight both terms. Below, we describe the training of the video-level classifier, the snippet classifier, and the snippet contrastive loss.
\textbf{Video Classifier Training.}
The video classifier is trained from a binary cross entropy loss to estimate if a video shows a polyp using the video-level labels. 
The loss $\ell_{vid}(.)$ from~\eqref{eq:main_loss} is the binary cross entropy loss defined as
\begin{equation}
  \ell_{vid}(\mathcal{D};\theta,\gamma) =
  -\sum_{(\mathbf{F}_i,y_i)\in\mathcal{D}}\big(y_i\log(v_{\gamma}(f_{\theta}(\mathbf{F}_i)) + (1-y_i)\log(1-v_{\gamma}(f_{\theta}(\mathbf{F}_i)))\big),
    \label{eq:BCE}
\end{equation}
where $v_{\gamma}:\mathcal{X} \to [0,1]$ is the video level anomaly classifier parameterised by $\gamma$.


\textbf{Snippet Classifier Training.}
The snippet classifier is optimised by training a top $k$ ranking loss function using a set that contains the $k$ snippets with the largest anomaly scores from $s_{\phi}(\mathbf{F})$ in~\eqref{eq:main_loss}.
More specifically, we propose the following loss $\ell_{snp}(.)$ from~\eqref{eq:main_loss} that maximises the separability between normal and abnormal videos:
\begin{equation}
   \ell_{snp}(\mathcal{D};\theta,\phi) = \sum_{\substack{(\mathbf{F}_i,y_i)\in\mathcal{D},y_i=1\\(\mathbf{F}_j,y_j)\in\mathcal{D},y_j=0}}\max \left(0, 1 - g_k(s_{\phi}(f_{\theta}(\mathbf{F}_i)) - g_k(s_{\phi}(f_{\theta}(\mathbf{F}_j)) )\right),
    \label{eq:L2C}
\end{equation}
where $g_k(.)$ returns the
mean anomaly score from $s_{\phi}(.)$ of the top $k$ snippets from a video~\cite{tian2021weakly,li2015multiple}. 
\textbf{Contrastive Snippet Mining.}
To make anomaly classification robust to hard normal and abnormal snippets, 
we propose the following novel snippet contrastive loss:
\begin{equation}
\ell_{cnt}(\mathcal{D};\theta) = \ell_{c}(\mathcal{D}^{HA},\mathcal{D}^{EA},\mathcal{D}^{EN};\theta) +
\ell_{c}(\mathcal{D}^{HN},\mathcal{D}^{EN},\mathcal{D}^{EA};\theta), 
    \label{eq:contrastive_loss}
\end{equation}
where $\mathcal{D}^{HA}$ and $\mathcal{D}^{EA}$ represent sets of hard and easy abnormal snippets, while $\mathcal{D}^{HN}$ and $\mathcal{D}^{EN}$ denote sets of hard and easy normal snippets,
\begin{equation}
\scalebox{0.85}{$
    \ell_{c}(\mathcal{D}^{HA},\mathcal{D}^{EA},\mathcal{D}^{EN};\theta) = \sum_{\mathbf{F}_i\in\mathcal{D}^{HA},\mathbf{F}_j\in\mathcal{D}^{EA}} 
\log\frac{\exp\left[ \frac{1}{\tau} {f_{\theta}(\mathbf{F}_{i})}^{\top} {f_{\theta}(\mathbf{F}}_{j}) \right]}{\exp\left[ \frac{1}{\tau} {f_{\theta}(\mathbf{F}_{i})}^{\top} {f_{\theta}(\mathbf{F}}_j) \right] + 
\sum_{\mathbf{F}_m\in\mathcal{D}^{EN}} \exp \left[ \frac{1}{\tau} {f_{\theta}(\mathbf{F}_{i})}^{\top} {f_{\theta}(\mathbf{F}}_{m})\right]},
$}
\end{equation}
and in a similar way we compute $\ell_{c}(\mathcal{D}^{HN},\mathcal{D}^{EN},\mathcal{D}^{EA};\theta)$.
The idea explored in~\eqref{eq:contrastive_loss} is to pull together easy and hard snippet features in $\mathcal{X}$ from the same class (normal or abnormal) and push apart features from different classes. 

\begin{wrapfigure}{ht}{.60\textwidth}

    \centering
    \includegraphics[width=0.5\textwidth]{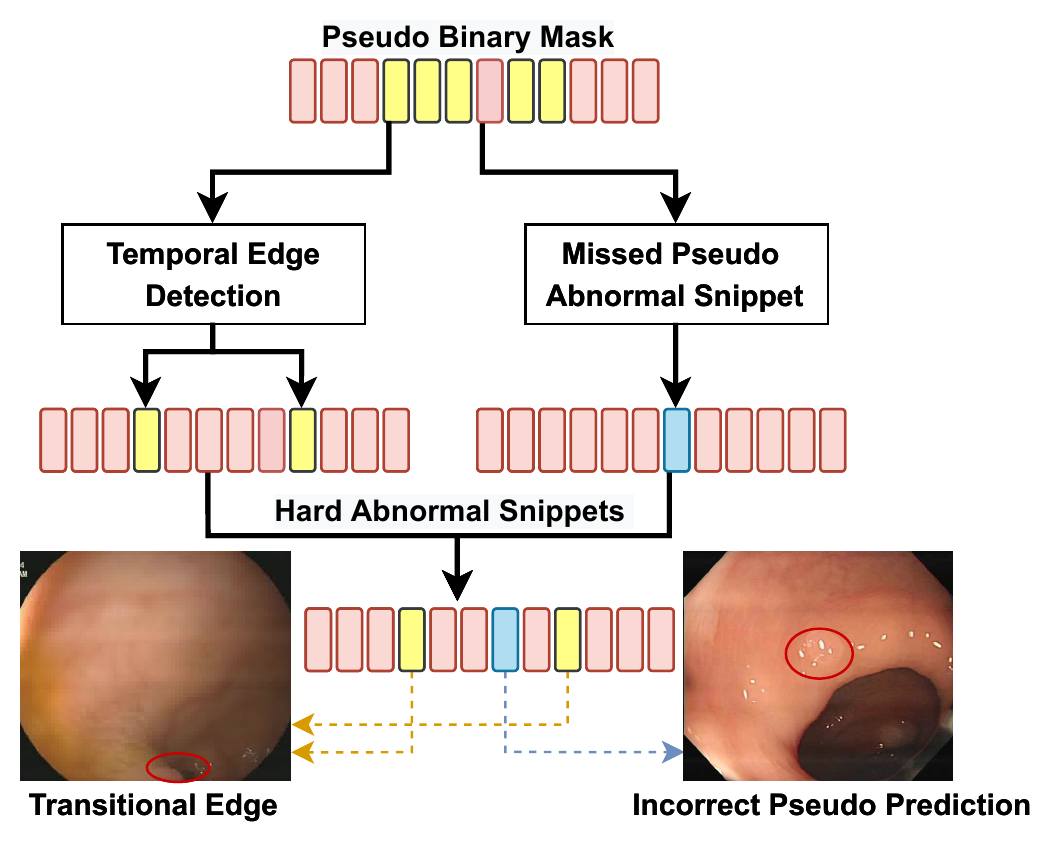}
    \caption{Hard abnormal snippet mining algorithm to select temporal edge snippets and missed pseudo abnormal snippets. Those two types of hard anomalies represent: 1) transitional frames where polyps may be partially visible, or 2) subtle (i.e., small and flat) polyps that can lead to incorrect low anomaly scores. 
    }
    \label{fig:hard_abn_mining}
    \vspace{-20pt}
\end{wrapfigure}

The selection of $\mathcal{D}^{HN},\mathcal{D}^{EN},\mathcal{D}^{HA},\mathcal{D}^{EA}$ and their incorporation into our MIL learning framework is one key contribution of this work to address the poor detection accuracy of hard anomalous snippets in existing WVAD methods.
Specifically, for abnormal videos, we first classify each of their $T$ snippets with $\hat{y}(t) = (\tilde{y}(t) > \epsilon)$, where $\tilde{y} = s_{\phi}(f_{\theta}(\mathbf{F}))$.
We then identify the temporal edge snippets and missed pseudo abnormal snippets as hard anomalies $\mathcal{D}^{HA}$. For temporal edge detection, we use the erosion operator
to subtract the original and eroded sequences and 
locate such transitional edge snippets, 
which are considered as hard anomalies (See Fig.~\ref{fig:hard_abn_mining} - temporal edge detection), and inserted into $\mathcal{D}^{HA}$.  
For locating the missed pseudo abnormal snippets, we assume that a subtle anomalous event (i.e., a small/flat polyp) happens in a region of $K$ consecutive snippets when $\frac{R}{K}$ (majority) of them have $\hat{y}(t) = 1$, where $K$ and $R$ are respectively the hyper-parameters to control the temporal length of the pseudo abnormal region and the ratio of the minimum number of the abnormal pseudo snippets inside that region. The incorrectly predicted normal snippets inside abnormal regions (i.e., missed abnormal snippets in Fig.~\ref{fig:hard_abn_mining}) are also inserted into $\mathcal{D}^{HA}$ as hard anomalies.

This hard anomaly selection process is motivated by the following two main observations: 1) subtle abnormal snippets from anomalous videos share similar characteristics to normal snippets (i.e., small and flat polyps) and consequently have low anomaly scores, and this can be easily identified from the adjacent abnormal snippets with higher anomaly scores since abnormal frames containing polyps are often contiguous; and 2) the transitional snippets between normal and abnormal events often contain noise such as water, endoscope pipe or partially visible polyps, so they are unreliable and can lead to inaccurate detection.  

Hard normal (HN) snippets (e.g., healthy frames containing water and feces) are collected by selecting the snippets with top $k$ anomaly scores from normal videos since normal videos do not have any abnormalities, so the ones with incorrectly predicted higher scores can be deemed as hard normal. 
For easy snippet mining, we hypothesise that the snippets with the smallest $k$ anomaly scores from normal videos and the snippets with top $k$ anomaly scores from abnormal videos are easy normal (EN) and easy abnormal (EA). 

\section{Experiments and Results}

\subsection{Dataset} To form a real-world large-scale video polyp detection dataset, we collected colonoscopy videos from two widely used public datasets: Hyper-Kvasir~\cite{borgli2020hyperkvasir} and LDPolypVideo~\cite{ma2021ldpolypvideo}. 
The new dataset contains 61 normal videos without polyps and 102 abnormal videos with polyps for training, and 30 normal videos and 60 abnormal videos for testing. The videos in the training set have video-level labels and the videos in testing set contain frame-level labels. 
This dataset contains over one million frames and has diverse polyps with various sizes and shapes, making it one of the largest and most challenging colonoscopy datasets in the field. 
The dataset setup will be publicly available upon paper acceptance.

\subsection{Implementation Details} 
Following~\cite{sultani2018real,tian2021weakly}, each video is divided into 32 video snippets, i.e., $T=32$.
For all experiments, we set $k = 3$ in~\eqref{eq:L2C}.
The 2048D input tokens are extracted from the '$mix\_5c$' layer of the pre-trained I3D~\cite{kay2017kinetics} network. Note that the I3D network is not fine-tuned on any medical dataset. For the transformer block, we set the number of heads to 8, depth of transformer blocks to 12, and use a 3 $\times $ 1 DW Conv1D. $\alpha$ and $\beta$ in \eqref{eq:loss_regularisation} are both set to $5e-4$. 
Our method is trained in an end-to-end manner using the Adam optimiser~\cite{kingma2014adam} with a weight decay of 0.0005 and a batch size of 32 for 200 epochs. 
The learning rate is set to 0.001. Following~\cite{sultani2018real,tian2021weakly}, each mini-batch consists of samples from 32 randomly selected normal and abnormal videos. The method is implemented in PyTorch~\cite{NEURIPS2019_9015} and trained with a NVIDIA 3090 GPU. The overall training times takes around 2.5 hours, and the mean inference time takes 0.06s per frame -- this time includes the I3D extraction time. 
For all baselines, we use the same I3D backbone and benchmark setup as ours.

\subsection{Evaluation on Polyp Frame Detection} 
\noindent\textbf{Baselines.}  We train six SOTA WVAD baselines:
DeepMIL~\cite{sultani2018real}, GCN-Ano~\cite{zhong2019graph}, CLAWS~\cite{zaheer2020claws}, AR-Net~\cite{wan2020arnet}, MIST~\cite{feng2021mist}, and RTFM~\cite{tian2021weakly}. The
same experimental setup as our approach is applied to these baselines for fair comparison.

\noindent\textbf{Evaluation Measures.} 
Similarly to previous papers~\cite{sultani2018real,gong2019memorizing}, we use the frame-level area under the ROC curve (AUC) as the evaluation measure. 
Given that the AUC can produce optimistic results for imbalanced problems, such as anomaly detection, we follow~\cite{Wu2020not,pang2021toward} and use average precision (AP) as another evaluation measure. 
Larger AUC and AP values indicate better performance.

\noindent\textbf{Quantitative Comparison.} We show the quantitative comparison results in Table~\ref{tab:main_result}. Our model achieves the best 98.4\% AUC and 86.6\% AP and outperforms all six SOTA
methods by a large margin. We obtain a maximum 10\% and a minimum 2\% AUC improvement, and a maximum 18\% and a minimum 6\% AP improvement over the second best approaches. Our method substantially surpasses the most recent WVAD approach RTFM~\cite{tian2021weakly} by 8\% AP.

\begin{table}[t!]
\centering
\resizebox{0.46\textwidth}{!}{%
\begin{tabular}{@{}c@{\hskip .15in} c @{\hskip .15in}  c @{\hskip .15in}  c@{}}
\toprule
Method & Publication & AUC & AP \\ \hline \hline
DeepMIL~\cite{sultani2018real} & CVPR'18 & 89.41 & 68.53 \\
GCN-Ano~\cite{zhong2019graph} & CVPR'19 & 92.13 & 75.39 \\
CLAWS~\cite{zaheer2020claws} & ECCV'20 & 95.62 & 80.42 \\
AR-Net~\cite{wan2020arnet} & ICME'20 & 88.59 & 71.58 \\
MIST~\cite{feng2021mist} & CVPR'21 & 94.53 & 72.85 \\
RTFM~\cite{tian2021weakly} & ICCV'21 & 96.30 & 77.96 \\ \hline
Ours &  & \textbf{98.41} & \textbf{86.63} \\ \bottomrule
	
\end{tabular}%
}
\caption{Comparison of frame-level AUC and AP performance with other SOTA WVADs on colonoscopy dataset using the same I3D feature extractor. }
\vspace{-20pt}
\label{tab:main_result}
\end{table}

\begin{figure}[t!]
    \centering
    \includegraphics[width=1\textwidth]{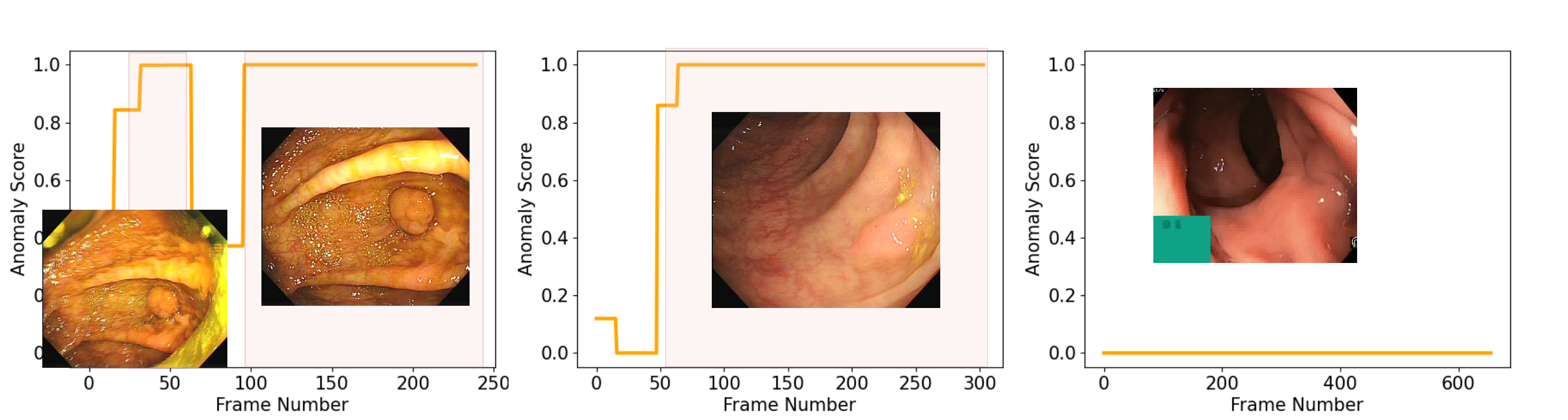}
    \caption{Anomaly scores (orange curve) of our method on test videos. Pink areas indicate the labelled testing abnormal events.  
    }
    \vspace{-10pt}
    \label{fig:qualitative}
\end{figure}

\noindent\textbf{Qualitative Comparison.} 
In Fig.~\ref{fig:qualitative}, we show the anomaly scores produced by our model for test videos from our polyp detection dataset. 
As illustrated by the orange curves, our model can effectively produce small anomaly scores for normal snippets and large anomaly scores for abnormal snippets. Our model is also able to detect multiple anomalous events (e.g., videos with two polyp event occurrences - first figure in Fig~\ref{fig:qualitative}) in one video. Also, our model can also detect the subtle polyps (middle figure in Fig~\ref{fig:qualitative}). 


\subsection{Ablation Study}

Tab.~\ref{tab:ablation} shows the contribution of each component of our proposed method on the testing set. The baseline top-k MIL network, trained with $\ell_{snp}$, achieves  92.8\% AUC and 71.9\% AP.
Our method obtains a significant performance gain by adding the proposed convolutional transformer encoder (CTE). Adding the video classifier, represented by the loss $\ell_{vid}(.)$, boosts the performance by about 2\% AUC and 3\% AP. The proposed hard/easy snippet contrastive loss, denoted by the loss $\ell_{cnt}(.)$, further improve the performance (e.g., increasing AP by about 4\%), indicating the effectiveness of addressing the hard anomaly issues.

\begin{table}[t!]
\centering
\scalebox{0.96}{%
\begin{tabular}{@{}@{\hskip .15in}c@{\hskip .15in}c@{\hskip .15in}c@{\hskip .15in}c@{\hskip .15in}|@{\hskip .15in}c@{\hskip .15in}c@{\hskip .15in}@{}}
\toprule
top-k ($\ell_{snp}$) & CTE    &  $\ell_{vid}$ & $\ell_{cnt}$ & AUC  & AP \\ \hline \hline
 \checkmark &    & &  & 92.88  & 71.96   \\
 \checkmark&    \checkmark  &  &    & 94.92 &	79.56   \\ 
 \checkmark &  \checkmark & \checkmark  &  & 96.74 & 82.88  \\ \hline
 \checkmark&    \checkmark & \checkmark & \checkmark & \textbf{98.41} & \textbf{86.63}
 \\ 
  \bottomrule
\end{tabular}%
}
\caption{Ablation studies for polyp frame detection. The linear network with top-k MIL ranking loss is considered as the baseline, and CTE denotes the Convolutional Transformer Encoder.   }
 \vspace{-20pt}
\label{tab:ablation}
\end{table}

\section{Conclusion}

We proposed a new transformer-based MIL framework as a robust anomaly classifier for detecting polyp frames in colonoscopy videos. 
To the best of our knowledge, our method is the first to formulate polyp detection as a weakly-supervised video anomaly detection problem, and also to introduce transformer to explore global temporal dependency between video snippets. We also proposed a novel and effective contrastive snippet mining (CSM) to enable an effective learning of challenging abnormal polyp frames (i.e., small and partially visible polyps) and normal frames (i.e., water and feces). The resulting anomaly classifier showed SOTA results on our proposed large-scale colonoscopy dataset. Despite the remarkable performance on detecting polyp frames, our model may fail for online inference due to the transformer self-attention operation. We plan to further investigate the online self-attention techniques in future work. 


%
%
%
\bibliographystyle{splncs04}
\bibliography{mybibliography}

\begin{thebibliography}{10}
\providecommand{\url}[1]{\texttt{#1}}
\providecommand{\urlprefix}{URL }
\providecommand{\doi}[1]{https://doi.org/#1}

\bibitem{arnab2021vivit}
Arnab, A., Dehghani, M., Heigold, G., Sun, C., Lu{\v{c}}i{\'c}, M., Schmid, C.:
  Vivit: A video vision transformer. In: Proceedings of the IEEE/CVF
  International Conference on Computer Vision. pp. 6836--6846 (2021)

\bibitem{borgli2020hyperkvasir}
Borgli, H., et~al.: Hyperkvasir, a comprehensive multi-class image and video
  dataset for gastrointestinal endoscopy. Scientific Data  \textbf{7}(1),
  1--14 (2020)

\bibitem{carreira2017quo}
Carreira, J., Zisserman, A.: Quo vadis, action recognition? a new model and the
  kinetics dataset. In: proceedings of the IEEE Conference on Computer Vision
  and Pattern Recognition. pp. 6299--6308 (2017)

\bibitem{chen2021deep}
Chen, Y., Tian, Y., Pang, G., Carneiro, G.: Deep one-class classification via
  interpolated gaussian descriptor. arXiv preprint arXiv:2101.10043  (2021)

\bibitem{chollet2017xception}
Chollet, F.: Xception: Deep learning with depthwise separable convolutions. In:
  Proceedings of the IEEE conference on computer vision and pattern
  recognition. pp. 1251--1258 (2017)

\bibitem{dosovitskiy2020image}
Dosovitskiy, A., Beyer, L., Kolesnikov, A., Weissenborn, D., Zhai, X.,
  Unterthiner, T., Dehghani, M., Minderer, M., Heigold, G., Gelly, S., et~al.:
  An image is worth 16x16 words: Transformers for image recognition at scale.
  arXiv preprint arXiv:2010.11929  (2020)

\bibitem{feng2021mist}
Feng, J.C., Hong, F.T., Zheng, W.S.: Mist: Multiple instance self-training
  framework for video anomaly detection. In: Proceedings of the IEEE/CVF
  Conference on Computer Vision and Pattern Recognition. pp. 14009--14018
  (2021)

\bibitem{gong2019memorizing}
Gong, D., et~al.: Memorizing normality to detect anomaly: Memory-augmented deep
  autoencoder for unsupervised anomaly detection. In: ICCV. pp. 1705--1714
  (2019)

\bibitem{ji2021progressively}
Ji, G.P., Chou, Y.C., Fan, D.P., Chen, G., Fu, H., Jha, D., Shao, L.:
  Progressively normalized self-attention network for video polyp segmentation.
  In: International Conference on Medical Image Computing and Computer-Assisted
  Intervention. pp. 142--152. Springer (2021)

\bibitem{kay2017kinetics}
Kay, W., Carreira, J., Simonyan, K., Zhang, B., Hillier, C., Vijayanarasimhan,
  S., Viola, F., Green, T., Back, T., Natsev, P., et~al.: The kinetics human
  action video dataset. arXiv preprint arXiv:1705.06950  (2017)

\bibitem{kingma2014adam}
Kingma, D.P., Ba, J.: Adam: A method for stochastic optimization. arXiv
  preprint arXiv:1412.6980  (2014)

\bibitem{li2015multiple}
Li, W., Vasconcelos, N.: Multiple instance learning for soft bags via top
  instances. In: Proceedings of the ieee conference on computer vision and
  pattern recognition. pp. 4277--4285 (2015)

\bibitem{liu2019photoshopping}
{Liu}, Y., {Tian}, Y., {Maicas}, G., {Cheng Tao Pu}, L.Z., {Singh}, R.,
  {Verjans}, J.W., {Carneiro}, G.: Photoshopping colonoscopy video frames. In:
  ISBI. pp.~1--5 (2020)

\bibitem{ma2021ldpolypvideo}
Ma, Y., Chen, X., Cheng, K., Li, Y., Sun, B.: Ldpolypvideo benchmark: A
  large-scale colonoscopy video dataset of diverse polyps. In: International
  Conference on Medical Image Computing and Computer-Assisted Intervention. pp.
  387--396. Springer (2021)

\bibitem{pang2021toward}
Pang, G., van~den Hengel, A., Shen, C., Cao, L.: Toward deep supervised anomaly
  detection: Reinforcement learning from partially labeled anomaly data. In:
  Proceedings of the 27th ACM SIGKDD Conference on Knowledge Discovery \& Data
  Mining. pp. 1298--1308 (2021)

\bibitem{pang2019deep}
Pang, G., Shen, C., van~den Hengel, A.: Deep anomaly detection with deviation
  networks. In: Proceedings of the 25th ACM SIGKDD International Conference on
  Knowledge Discovery \& Data Mining. pp. 353--362 (2019)

\bibitem{NEURIPS2019_9015}
Paszke, A., Gross, S., Massa, F., Lerer, A., Bradbury, J., Chanan, G., Killeen,
  T., Lin, Z., Gimelshein, N., Antiga, L., Desmaison, A., Kopf, A., Yang, E.,
  DeVito, Z., Raison, M., Tejani, A., Chilamkurthy, S., Steiner, B., Fang, L.,
  Bai, J., Chintala, S.: Pytorch: An imperative style, high-performance deep
  learning library. In: Wallach, H., Larochelle, H., Beygelzimer, A.,
  d~Alch\'{e}-Buc, F., Fox, E., Garnett, R. (eds.) Advances in Neural
  Information Processing Systems 32, pp. 8024--8035. Curran Associates, Inc.
  (2019),
  \url{http://papers.neurips.cc/paper/9015-pytorch-an-imperative-style-high-performance-deep-learning-library.pdf}

\bibitem{pu2020computer}
Pu, L.Z.C.T., et~al.: Computer-aided diagnosis for characterisation of
  colorectal lesions: a comprehensive software including serrated lesions.
  Gastrointestinal Endoscopy  (2020)

\bibitem{sultani2018real}
Sultani, W., Chen, C., Shah, M.: Real-world anomaly detection in surveillance
  videos. In: Proceedings of the IEEE Conference on Computer Vision and Pattern
  Recognition. pp. 6479--6488 (2018)

\bibitem{tian2021self}
Tian, Y., Liu, F., et~al.: Self-supervised multi-class pre-training for
  unsupervised anomaly detection and segmentation in medical images. arXiv
  preprint arXiv:2109.01303  (2021)

\bibitem{tian2020few}
Tian, Y., Maicas, G., Pu, L.Z.C.T., Singh, R., Verjans, J.W., Carneiro, G.:
  Few-shot anomaly detection for polyp frames from colonoscopy. In:
  International Conference on Medical Image Computing and Computer-Assisted
  Intervention. pp. 274--284. Springer (2020)

\bibitem{tian2021detecting}
Tian, Y., otherss: Detecting, localising and classifying polyps from
  colonoscopy videos using deep learning. arXiv preprint arXiv:2101.03285
  (2021)

\bibitem{tian2021weakly}
Tian, Y., Pang, G., Chen, Y., Singh, R., Verjans, J.W., Carneiro, G.:
  Weakly-supervised video anomaly detection with robust temporal feature
  magnitude learning. In: Proceedings of the IEEE/CVF International Conference
  on Computer Vision. pp. 4975--4986 (2021)

\bibitem{tian2021constrained}
Tian, Y., Pang, G., Liu, F., Shin, S.H., Verjans, J.W., Singh, R., Carneiro,
  G., et~al.: Constrained contrastive distribution learning for unsupervised
  anomaly detection and localisation in medical images. MICCAI 2021  (2021)

\bibitem{tian2019one}
Tian, Y., et~al.: One-stage five-class polyp detection and classification. In:
  2019 IEEE 16th International Symposium on Biomedical Imaging (ISBI 2019). pp.
  70--73. IEEE (2019)

\bibitem{wan2020arnet}
{Wan}, B., {Fang}, Y., {Xia}, X., {Mei}, J.: Weakly supervised video anomaly
  detection via center-guided discriminative learning. In: 2020 IEEE
  International Conference on Multimedia and Expo (ICME). pp.~1--6 (2020)

\bibitem{wu2021cvt}
Wu, H., Xiao, B., Codella, N., Liu, M., Dai, X., Yuan, L., Zhang, L.: Cvt:
  Introducing convolutions to vision transformers. arXiv preprint
  arXiv:2103.15808  (2021)

\bibitem{Wu2020not}
Wu, P., Liu, j., Shi, Y., Sun, Y., Shao, F., Wu, Z., Yang, Z.: Not only look,
  but also listen: Learning multimodal violence detection under weak
  supervision. In: European Conference on Computer Vision (ECCV) (2020)

\bibitem{zaheer2020claws}
Zaheer, M.Z., Mahmood, A., Astrid, M., Lee, S.I.: Claws: Clustering assisted
  weakly supervised learning with normalcy suppression for anomalous event
  detection. In: European Conference on Computer Vision. pp. 358--376. Springer
  (2020)

\bibitem{zhong2019graph}
Zhong, J.X., Li, N., Kong, W., Liu, S., Li, T.H., Li, G.: Graph convolutional
  label noise cleaner: Train a plug-and-play action classifier for anomaly
  detection. In: Proceedings of the IEEE Conference on Computer Vision and
  Pattern Recognition. pp. 1237--1246 (2019)

\end{thebibliography}
%




\end{document}